\documentclass[letterpaper]{article}

\usepackage{natbib,alifeconf}  
\usepackage{url,hyperref,cleveref}
\usepackage{booktabs}
\usepackage{lipsum}

\title{
Morphological Cognition: Classifying MNIST Digits Through Morphological Computation Alone
}

\author{
   Alican Mertan$^{1}$, \and
   Nick Cheney$^{1}$ \\
   \mbox{}\\
   $^1$Neurobotics Lab, University of Vermont, USA \\
   \{alican.mertan, ncheney\}@uvm.edu
} 

\begin{document}

\maketitle

\begin{abstract}
    With the rise of modern deep learning, neural networks have become an essential part of virtually every artificial intelligence system, making it difficult even to imagine different models for intelligent behavior. In contrast, nature provides us with many different mechanisms for intelligent behavior, most of which we have yet to replicate. One of such underinvestigated aspects of intelligence is embodiment and the role it plays in intelligent behavior. In this work, we focus on how the simple and fixed behavior of constituent parts of a simulated physical body can result in an emergent behavior that can be classified as cognitive by an outside observer. Specifically, we show how simulated voxels with fixed behaviors can be combined to create a robot such that, when presented with an image of an MNIST digit zero, it moves towards the left; and when it is presented with an image of an MNIST digit one, it moves towards the right. Such robots possess what we refer to as ``morphological cognition'' -- the ability to perform cognitive behavior as a result of morphological processes. To the best of our knowledge, this is the first demonstration of a high-level mental faculty such as image classification performed by a robot without any neural circuitry. We hope that this work serves as a proof-of-concept and fosters further research into different models of intelligence.
\end{abstract}

\section{Introduction}

Artificial intelligence (AI) broadly refers to the replication of intelligence, whether it is human-like or conforms to any other definition of intelligence, in different media such as in machines~\citep{russell2016artificial} or out of living tissue~\citep{kriegman_scalable_2020}. Although intelligence itself is a complex, multi-faceted phenomenon that is elusive to definitions~\citep{legg2007collection}, the study of modern AI has become synonymous with the study of deep learning, the root of its most recent success~\citep{lecun_deep_2015}.

\paragraph{An old bias} 
This particular branch of AI, deep learning, mainly refers to the study of training, understanding, and employing artificial neural networks to tackle a broad class of problems. Artificial neural networks are initially conceptualized to study the ``calculus of neural activity'', with strong biological plausibility~\citep{mcculloch1943logical}. Even though the biological plausibility of neural networks seems to be less sought after nowadays, the way neural networks are employed still conceptually resembles brains. And we have a bias toward attributing intelligent behavior solely as a result of processes confined in the brain, a bias that is far from new, dating back to Descartes' mind-body dualism~\citep{descartes_discourse_1993}. In fact, Lamarck takes this to the extreme: ``No mental function shall be ascribed to an organism for which the complexity of the nervous system of the organism is insufficient''~\citep{Bateson2002}.

\paragraph{Embodied cognition approach} 
Nature, on the other hand, provides many examples of intelligent behavior stemming from non-neural mechanisms. From macro scales, foraging, communication, memory, and learning behaviors in plants~\citep{trewavas2003aspects,karban2008plant}, to micro scales, predation, swarming, and social behaviors in prokaryotes~\citep{lyon_reframing_2021,lyon_basal_2023}, examples of complex, intelligent-seeming behaviors stemming from non-neural processes abound in nature.

Within the field of AI, evolutionary robotics (ER)~\citep{nolfi_evolutionary_2000} emerges as the branch that investigates the interplay between many factors as the driving force of intelligent behavior. ER started in the early 1990s, as a response to the shift of the then mainstream AI literature from creating lifelike intelligence to achieving ``efficient signal processing, optimal control, and data mining''~\citep{floreano2008bio}. As an approach to creating intelligence inspired by biological intelligence, ER places emphasis on experimenting with embodied and situated agents and how they interact with their environments, looking for different sources for creating intelligent behaviors.

\paragraph{What is missing?}
One particular source of intelligence is of main interest in this work, namely the body of an agent. How the body creates or affects how intelligent behavior arises is investigated in ER~\citep{pfeifer2006body}. Early work combined fixed bodies with separately postulated simple neural networks to create agents that achieve cognitive functions~\citep{beer_biological_1990,parisi1990econets,harvey1994seeing,beer1996toward,tuci_active_2009}. 
The following work investigates how the form of the morphology creates, affects, or facilitates behavior and how it can be optimized~\citep{sims_evolving_1994,hallam_method_2002,bongard_resilient_2006}. 
Later work provides the theoretical foundation for the inherent power of analog computation in physical systems~\citep{hauser2011towards}, and many works~\citep{nakajima_soft_2013,nakajima_information_2015,eder_morphological_2018,hauser_leveraging_2023} build on this theoretical basis and implement systems to utilize the computation in physical systems. In these works, the passive dynamics of the morphology are exploited as a physical reservoir to carry out some computation in aid of an abstract control layer~\citep{hauser_morphological_2014}. 
One important aspect of most work, including mainstream robotics work, a limitation as we will argue, is the treatment of the body without any inherent functions. The robot bodies are embodied in a physical environment, and the dynamics resulting from physical interactions with the environment are exploited, but the bodies do not have any inherent behavior -- they are under the control of an abstract control layer.
Although ER initially set out to achieve lifelike intelligence by studying all sources of intelligence, we suspect that ``the unreasonable effectiveness of deep learning''~\citep{sejnowski_unreasonable_2020} overshadowed the investigation of what bodies mean for intelligence, especially how they can be a source of intelligent behavior, as opposed to passively participating in its display. 

This is in stark contrast to nature -- natural evolution works with functional, active, and agential materials on every scale of organisms to create intelligent behavior~\citep{levin_darwins_2023}, which is considered an important aspect of natural evolution, and its implications for artificial evolution and AI should be investigated~\citep{bongard_theres_2023,hartl_evolutionary_2024}. It is hypothesized that the robustness, generality, and adaptation of natural intelligence could emerge from interactions of different scales of competency. On this theoretical basis, we argue that ER should investigate working with active materials, where bodies consist of parts with autonomous behaviors, to realize lifelike intelligence.

Moreover, there are practical challenges that arise from combining passive morphologies with separately postulated controllers modeled by neural networks. Recent literature shows that controllers optimized for particular morphologies become overly specialized and do not transfer well to other morphologies, commonly referred to as fragile co-adaptation~\citep{cheney_difficulty_2016,mertan_modular_2023}, and this phenomenon hinders evolutionary algorithms' ability to optimize brains and bodies together as complete agents~\citep{mertan_investigating_2024,mertan_controller_2025}. It has been argued, and we claim here as well, that bodies that are not passive and under complete control of a brain but rather enriched with autonomous functions and behaviors can alleviate fragile co-adaptation and enable better brain-body co-optimization. On this practical basis, we investigate the use of morphologies that consist of active parts to achieve optimization of complete agents that can exploit all potential sources of intelligence.

\paragraph{Morphological Cognition} Building on these ideas, we suggest not only examining morphology as an aid to the brain, but also studying morphology as doing full cognitive behavior end-to-end. We term robots that demonstrate cognitive behaviors without an abstract control layer as possessing ``morphological cognition''. In this work, we investigate the incorporation of more complex materials, ones that have fixed and simple, yet active and responsive behaviors, to increase the computational abilities of the resulting robots to achieve intelligent behavior. In the simulated voxel-based soft robotics setting, we experiment with combining ``smart'' voxels, such as ones that expand and contract continuously and ones that can expand and contract in response to a simple stimulus from the environment.

These materials are similar to what was previously proposed in the literature. \citet{cheney_unshackling_2014} evolved soft robots that consist of materials that constantly expand and contract. The design of robots with such voxels is optimized with evolution, creating robots that exhibit locomotion ability. Similar robot designs are fabricated in vivo using living cells~\citep{kriegman_scalable_2020}. Following a similar methodology, \citet{corucci_evolving_2016} evolved swimming soft robots. \citet{cheney2014evolved} evolves electrophysiological robots that can locomote under the electrical signal that flows through their voxels. 
These robots have only evolved for locomotion and lack the ability to respond to their environment in an intelligent way.

The most similar to this work is the work of \citet{corucci_material_2016} and \citet{mertan2024no}. The former evolved stationary robots to reach a light source, and the latter evolved soft robots that can adapt their behavior in response to their environment using voxels that can respond to stimuli by expanding or contracting. We build on this work by increasing the complexity of the stimulus to demonstrate that the combination of these materials is enough to achieve more complex tasks that can be categorized as classification.

\paragraph{Contribution}
We replicate a cognitive faculty, namely image classification, in simulated soft robots that lack any neural, or otherwise any type of, controller. We demonstrate that high-level intelligent behavior can arise from basic shape-transforming processes. Specifically, we evolve soft robots that consist of voxels with fixed and simple behaviors to classify MNIST digits~\citep{deng2012mnist}, that is, robots move toward the left side of the screen if the image of a digit they sense is one class, and they move toward the right side of the screen if the image of a digit they sense is another class. 
To the best of our knowledge, this is the first demonstration of a high-level cognitive ability, such as image classification, achieved in artificial creatures devoid of any analogy to neurons or any abstract control layer.

\begin{figure*}
    \centering
    \includegraphics[width=\linewidth]{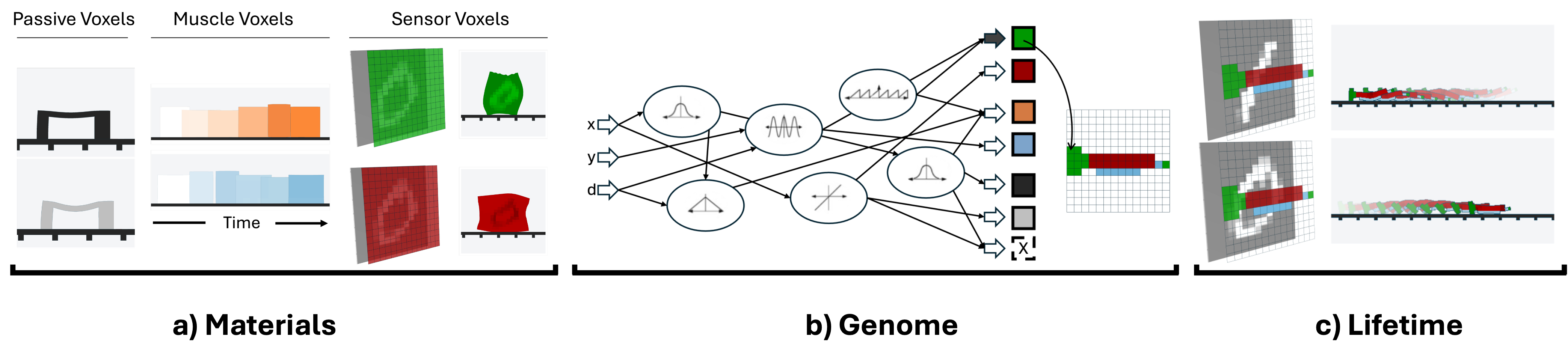}
    \caption{Overview of the experimental setting. (a) The type of voxels available to evolution and their behaviors. Passive voxels (black and gray) are used to support the structure of the robot's morphology. Muscle voxels (orange and teal) continuously expand and contract to provide energy for locomotion. Sensor voxels (red and green) change their area based on the intensity of the pixel that they correspond to. The shade of color of each voxel represents the area of the voxel, where darker is smaller. Please note how the MNIST images are clearly visible on robots made out of entirely sensor voxels. (b) Robot morphologies are encoded by compositional pattern producing networks. Evolution alters these networks to discover designs with the desired behavior. (c) Robots are evaluated under the presence of images of different MNIST digits and are evolved to locomote in different directions in response to digits from different classes.}
    \label{fig:teaser}
\end{figure*}

\section{Methodology}

\paragraph{Problem description}
We are interested in replicating classification ability in simulated voxel-based soft robots consisting of parts with simple and fixed behavior. Specifically, we choose the classification of MNIST digits, a commonly used benchmark for machine learning algorithms~\citep{baldominos2019survey}. We use evolutionary optimization algorithms to optimize the design of robots to achieve the desired behavior of walking in opposite directions in response to images of MNIST digits from different classes.  
Here, we take a behaviorist perspective in attributing the classification ability to robots; that is, if a robot moves in different directions in response to different stimuli, it can be said that the robot is capable of classifying different stimuli.
The rest of this section explains our experimental setting in detail. For an overview of the system, see Fig.~\ref{fig:teaser}.

\paragraph{Simulation}
We run our experiments in Evogym~\citep{bhatia_evolution_2021}, a 2D voxel-based soft body simulator, where the robots are modeled by a spring-mass system. 
Each voxel consists of four point masses and six springs connecting them (four of them connect the point masses along the edges of a rectangle, two of them connect cross faces). 
The properties of these springs determine how rigid or soft the voxel is. The voxels can be actuated by changing the resting length of these springs, which results in a change in their area. 
Evogym provides four basic types of voxels. There are two passive voxels, rigid and soft. Passive voxels do not actuate by changing the resting length of their springs; rather, they respond to external forces from surrounding voxels and the environment. Lastly, there are two types of active voxels: horizontally actuating and vertically actuating active voxels. Active voxels can actuate by changing the resting length of their springs, as well as responding to external forces. 

The soft robots are essentially a collection of voxels in a 2D grid. For simplicity, we assume neighboring voxels are connected, which means that they share the touching point masses and springs. Designing a robot simply means deciding what type of voxel is going to be in each grid location (or lack of a voxel) in a given bounding box.

\paragraph{Environment}
We use the Python API exposed by the developers of Evogym~\citep{bhatia_evolution_2021} to create our custom environment. 
Initially, we determine which voxel types will be accessible for the evolutionary processes. Fig.~\ref{fig:teaser} (a) illustrates the types of available voxels and their behaviors. First, we keep the rigid (black) and soft (gray) passive voxels unchanged, allowing evolution to include passive voxels to support the structure of the robots. 

Next, we implement new voxel types out of the basic types provided by the simulator. For simplicity, all of the new voxels are implemented using the horizontally actuating basic voxel type.

To provide energy for locomotion, we create a new voxel type called muscle voxel, which expands and contracts continuously following a sinusoidal signal. The muscle voxel has two variants, one that actuates in phase (orange) and one that actuates out of phase (teal). This type of voxel is similar to the one used in~\cite{cheney_unshackling_2014,corucci_evolving_2016,mertan2024no}. Moreover, living cells with similar behavior are used to create computer-designed locomoting living organisms~\citep{kriegman_scalable_2020}.

To allow robots to \textit{see} images of MNIST digits, we create a new type of voxel, the sensor voxel. Sensor voxel has the behavior of changing its area proportionally to the intensity of a pixel in the presented image. There are two variants of sensor voxels, one that expands as the intensity of the pixel increases (green) and one that shrinks (red). This type of voxel is similar to the one used in~\cite{corucci_material_2016,mertan2024no}.
The pixel to which each sensor voxel responds is determined based on the placement of the sensor voxel in the robot's body. We assume a one-to-one correspondence between voxels and pixels, as illustrated in Fig.~\ref{fig:teaser} (c). This correspondence is not affected by the movement of the robot; that is, the sensor voxels always respond to the same corresponding pixel in the image. 

\paragraph{Evolutionary Algorithm}
We use the MAP-Elites algorithm~\citep{mouret_illuminating_2015} to evolve a diverse set of robots. The MAP-Elites algorithm explicitly maintains a diverse set of solutions, where the diversity is defined by the experimenter. We define three dimensions of diversity based on morphological attributes, namely the number of active, sensor, and total voxels. 
Each of these attributes is discretized into seven bins, resulting in an archive of size 90.
The diversity in these attributes allows the evolutionary algorithm to explore different morphologies, potentially discovering entirely different mechanisms for classification.

\paragraph{Evaluation and Selection}
The simulation's 2D design restricts robots to one-dimensional movement, enabling only binary classification; thus, we group the digits into two clusters. 
For instance, robots are evolved to walk in one direction when presented with images of digit zero, and to walk in the opposite direction when presented with images of digit one, as shown in Fig.~\ref{fig:teaser} (c).

The robots are evaluated for ten actuation cycles per image shown. The change in the center of mass of the robot is calculated to determine the horizontal movement of the robot in response to the presented images. If the robot is successfully walking in opposite directions for different digits, the minimum horizontal distance traveled in response to any of the presented images is used as the robot's fitness. If the robot is walking in the same direction for some digits, the maximum horizontal distance traveled in response to misclassified images is multiplied by $-1$ and used as fitness. In essence, fitness encourages the robot to move in opposite directions for different digits as fast as possible, without enforcing a particular direction for any digit.

\paragraph{Genome}
Robots are indirectly encoded by compositional pattern producing networks (CPPNs)~\citep{stanley_compositional_2007}. CPPN is a directed acyclic graph, where each node represents a mathematical function. 
The edges represent the weighted transfer of the output from one node to another as input.
If a node has multiple incoming edges, the transferred values are summed. Here we use the sine, the absolute value, the square, the square root, and their negatives ($-1 \times f(.)$) as activation functions within each node. 

Fig.~\ref{fig:teaser} (b) sketches how CPPNs are used to create robots. Each CPPN has three input nodes, namely the coordinates of a voxel ($(x,y)$) and its distance to the center of the grid, and 7 output nodes, 6 of them representing different types of voxels and one of them representing the lack of a voxel. The genotype-to-phenotype mapping is performed by querying the CPPN for each grid location and assigning the type of voxel indicated by the output node with the maximum value to that grid location.

During evolution, offspring are created by mutating the CPPN. The mutation operator randomly selects one of the following actions with equal probability: it can add or remove nodes or edges, modify a node's activation function, or alter an edge's weight.

\paragraph{Parameters}
Due to limited computational resources, we experiment with robots that are at most 14-by-14 voxels. To get the one-to-one correspondence between voxels and pixels, we downscale MNIST images to 14-by-14 by removing every other row and column.  

For all experiments, the evolutionary algorithm is run for 30,000 generations. 20 randomly selected individuals from the archive produce offspring at each generation through mutation. Each experiment is repeated five times with different random seeds. The code to reproduce experiments and supplementary materials are available \href{https://github.com/mertan-a/morphological-cognition}{here}.

\begin{figure*}[t]
    \centering
    \includegraphics[width=0.9\linewidth]{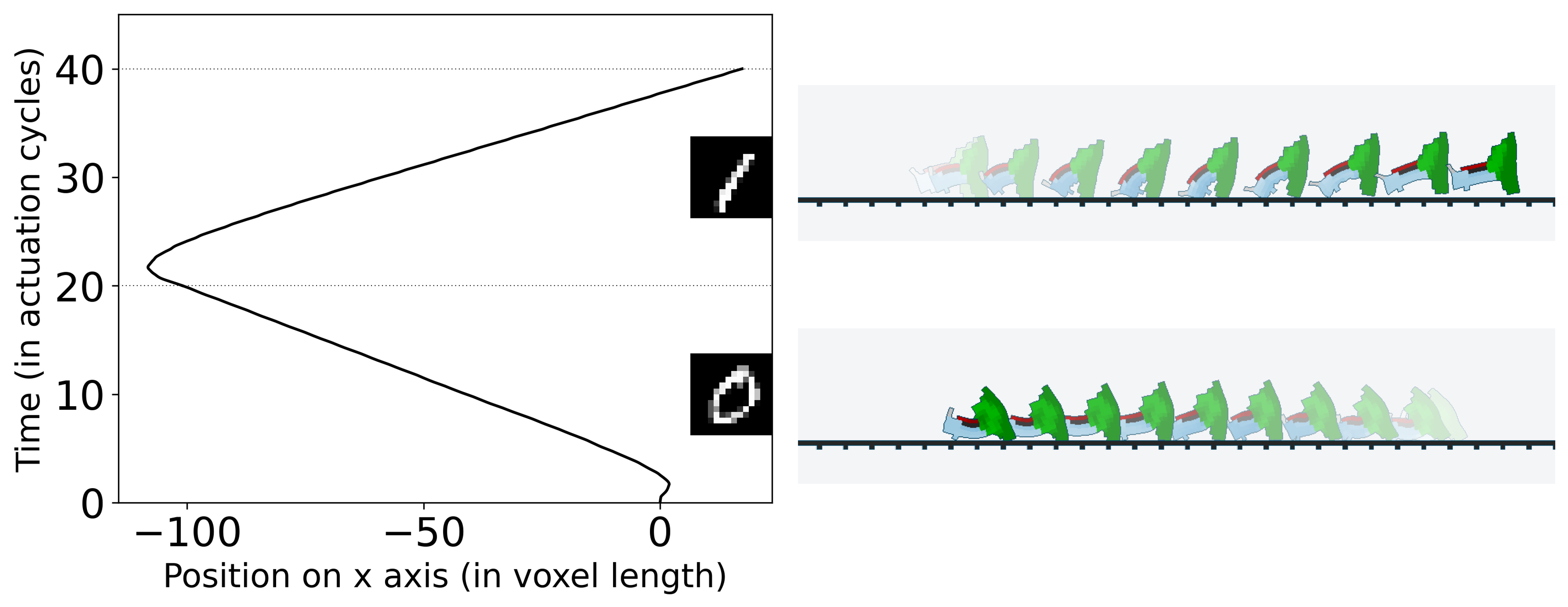}
    \caption{Space-time diagram of a selected individual (left) and snapshots of the individual's behavior under different stimuli superimposed on top of each other (right). The dashed lines mark the times when the stimuli change, and the stimulus presented to the robot is shown in the inset images. 
    The robot employs all three types of voxels, namely passive, muscle, and sensor voxels, to exhibit different gaits under different stimuli. The sensor voxels are placed to pay attention mostly right side of the image. The muscle voxels create an \textit{appendage}, which is used to pull the rest of the body toward the left side of the screen when the robot is presented with an image of zero, and it is used to push the body toward the right side of the screen when the robot is presented with an image of one. The way this appendage is used is determined by the sensor voxels, depending on the sensed image. }
    \label{fig:zero-one-std-bhc}
\end{figure*}

\section{Classifying MNIST Digits}

\subsection{Zero-One Classification}
First, we evolve robots made out of simple voxels to classify the MNIST digits of zeros and ones. To keep the computational cost tractable, we use a specific example of a zero and a one to evaluate the fitness of each individual. This can be seen as analogous to training a classifier with a single pair of examples in machine learning. 

\begin{figure*}[t]
    \centering
    \includegraphics[width=0.8\linewidth]{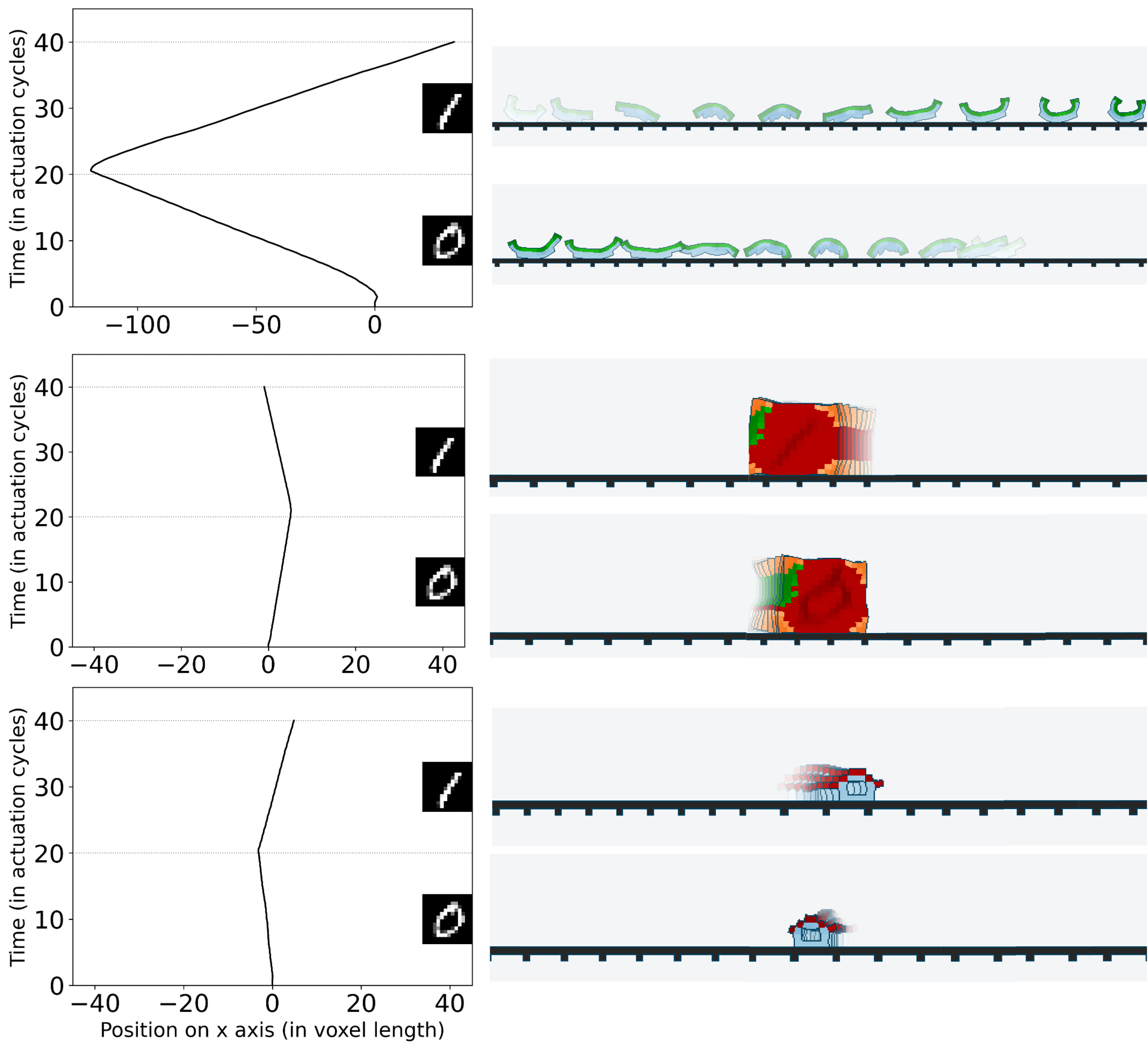}
    \caption{Space-time diagram of selected individuals (left) and snapshots of the individuals' behavior (right). 
    (Top) Evolution discovers that sensing a horizontal strip across the image is enough to alter the behavior of the robot and change its direction of locomotion. Intuitively, only paying attention to a horizontal strip is enough to recognize the zeros and ones.
    (Middle) A robot primarily constructed from sensor voxels has constrained speed due to a shortage of muscle voxels. The homogeneous patch of sensor voxels shows what the robot senses. 
    (Bottom) A robot with minimal sensor voxels. Evolution discovers that only six sensor voxels are enough to change the direction of locomotion.
    }
    \label{fig:zero-one-std-bhc-stacked}
\end{figure*}

Fig.~\ref{fig:zero-one-std-bhc} shows the space-time diagram of a selected individual (left) and its behavior under different stimuli is visualized as superimposed snapshots from the simulation engine (right). 
The robot consists of all three types of voxels, namely passive, muscle, and sensor voxels. The optimized design creates the emergent behavior in which the robot uses the left part of its body as an \textit{appendage}. This appendage consists of muscle voxels and is used to pull the body toward the left side of the screen when the robot senses an image of zero. When the robot is presented with an image of one, changes in the sensor voxels' area alter the behavior of the robot, and the appendage is used to push the robot toward the right side of the screen. 
From a behaviorist standpoint, this significant shift in behavior when exposed to various stimuli indicates that the robot possesses the ability to identify distinct digits.

Fig.~\ref{fig:zero-one-std-bhc-stacked} shows three interesting morphologies discovered by evolution. Fig.~\ref{fig:zero-one-std-bhc-stacked} (top) shows a robot that only senses a horizontal strip across images to alter its behavior. Intuitively, a horizontal strip is enough to recognize whether an image is one or zero. Fig.~\ref{fig:zero-one-std-bhc-stacked} (middle) and (bottom) show two robots that are capable of recognizing the two digits in very different ways. The middle robot consists mostly of sensor voxels. The homogeneous patch of sensor voxels displays what the robot senses. On the other hand, the bottom robot only uses six sensor voxels. Evolution discovered that paying attention to only six pixels is enough to change the relative position of the robot's center of mass, changing its direction of locomotion. The morphology of the robots directly exposes what the robots are paying attention to. 

\begin{figure*}[t]
    \centering
    \includegraphics[width=\linewidth]{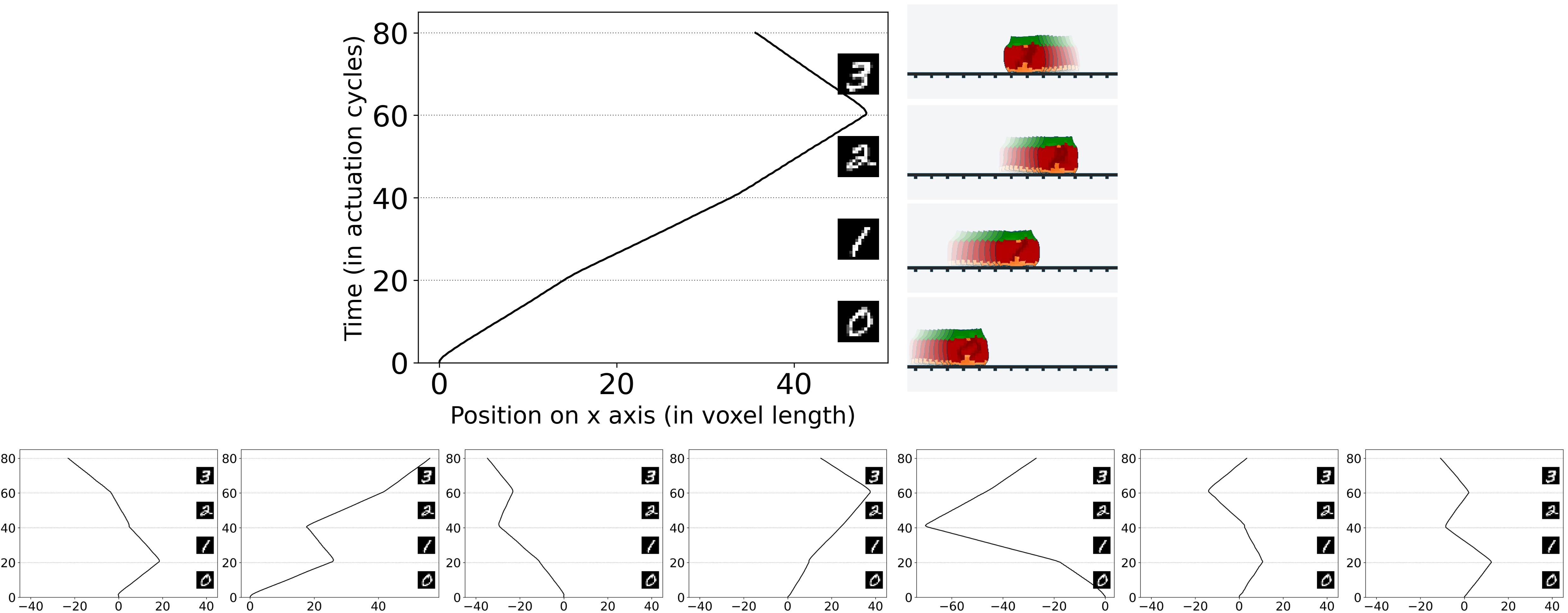}
    \caption{
    (Top) Space-time diagram of a robot and its behavior as superimposed snapshots during its lifetime. The robot moves toward the right side of the screen in response to all digits but digit three. The sensed digit is clearly ``visible'' on the homogeneous patch of sensor voxels.
    (Bottom) Space-time diagrams of selected robots performing all available tasks with four digits. The first four space-time diagrams depict the behavior of robots that move in one direction in response to images of all digits but one, for which it moves in the other direction. The last three space-time diagrams depict the behavior of robots that move in one direction in response to images of two of the digits and in the opposite direction in response to the images of the other two digits.
    In a single evolutionary run, evolution discovers different robots performing challenging cognitive tasks.
    }
    \label{fig:four-digit-all}
\end{figure*}

\subsection{4-digit classification}
To increase the challenge, we conduct an experiment in which robots evolved to classify four MNIST digits. Since robots are capable of moving in one dimension, we expect the robots to divide the digits into two groups. This setting gives us seven different classification tasks, four of them where robots are moving in the same direction in response to all digits but one, and three of them where robots are moving in opposite directions in response to two digits.
We combine the morphological diversity dimensions in the MAP-Elites' archive with a behavioral dimension representing these seven tasks to evolve robots that perform all tasks in a single evolutionary run. 
The fitness evaluations are done with a single example for each digit class.

Fig.~\ref{fig:four-digit-all} (top) shows a selected robot's space-time diagram and snapshots from its lifetime under different stimuli from simulation. The robot moves toward the right side of the screen in response to images of zero, one, and two, but when it senses an image of three, it changes its direction of locomotion and starts moving toward the left side of the screen. This change is purely a result of the morphological changes caused by the sensor voxels, which are ``visible'' on the homogeneous patch of sensor voxels.

Fig.~\ref{fig:four-digit-all} (bottom) shows the space-time diagram of selected individuals performing all available tasks with four digits. The robots in the first four space-time diagrams show a preference for a single digit out of the four, moving in one direction when presented with an image of all digits but one. The robots in the last three space-time diagrams cluster the digits into two bins. They move in one direction when they sense an image belonging to two of the digits, and move in the opposite direction when they sense an image belonging to the other two digits. Evolution is capable of finding robots with all the desired behaviors only by combining voxels with simple and fixed behaviors. 

\subsection{Generalization and morphological cognition capacity}

In all of our experiments, we use a randomly selected representative image for each experimented class from MNIST. This is analogous to training a machine learning system with a training set consisting of a single example from each class\footnote{Please note that this approach differs from the typical definition of single-shot learning in machine learning, where a \textit{pretrained} classifier is used to identify instances of an unseen class from a single labeled example.}. In this section, we investigate how these robots generalize outside of their training set. 

Some of the individuals in the final archive exploit the simulation engine's inaccuracies to achieve the desired behavior. 
Instead of rerunning experiments with carefully tuned fitness functions to eliminate such solutions, we decided to perform a qualitative selection after evolution to eliminate solutions that exploit the simulation engine.
We then randomly select ten individuals for this analysis to keep the computational costs tractable. Selected individuals are presented with 1000 images evenly divided between the experimented MNIST classes. For each image, the simulation is run for 10 actuation cycles, and the change in the center of mass is recorded. If the robot does not move more than three voxels in length, that image is counted as misclassified. The sign of the robot's center of mass's x-component is used to determine the classification made by the robot.

\begin{figure}[t]
    \centering
    \includegraphics[width=\linewidth]{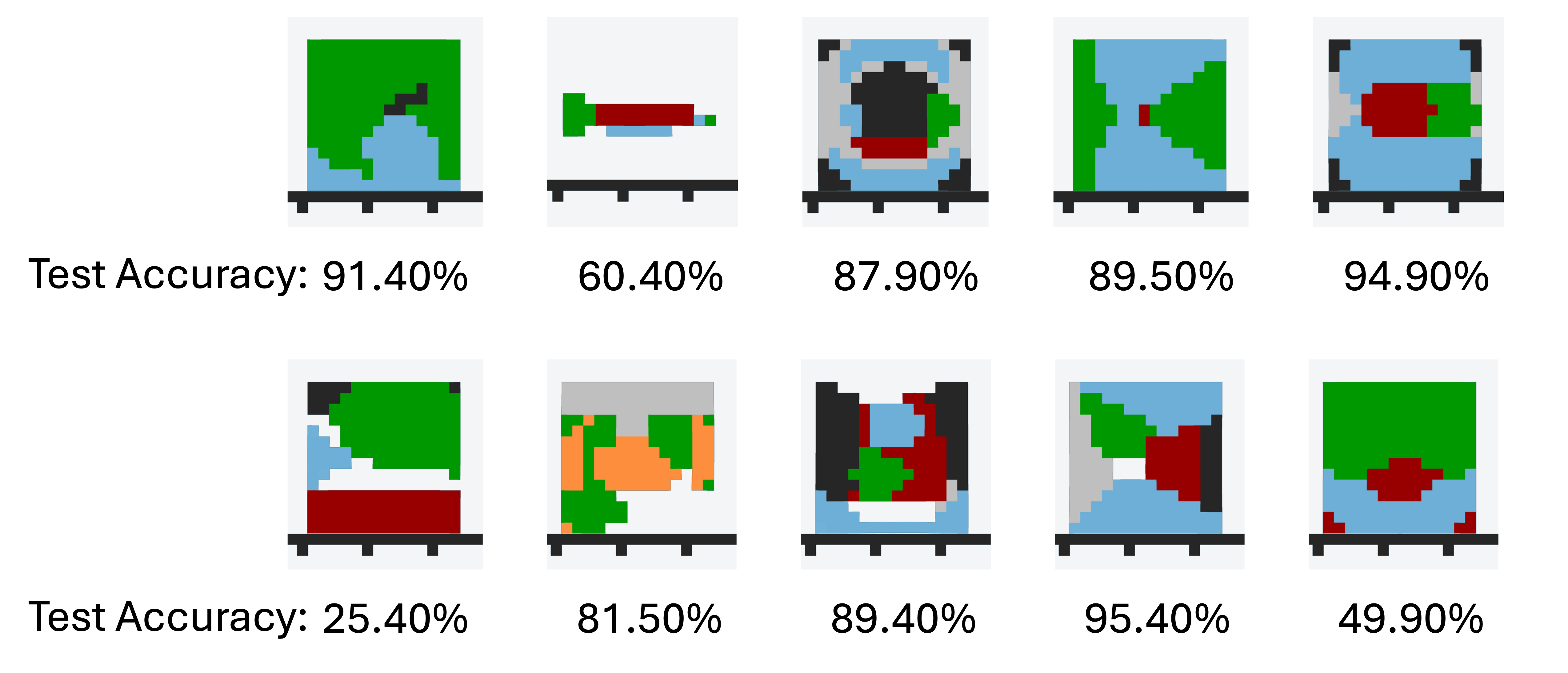}
    \caption{10 randomly selected robots that are evolved to recognize images of zero and one, and their performances on 1000 held-out MNIST images. The average accuracy on this set of robots is 76.57\%. However, performances vary a lot. 
    Considering that these robots have evolved to recognize a particular image of a zero and a one, and their performance is never evaluated on any other image during evolution, they show remarkable generalization ability. }
    \label{fig:zero-one-generalization}
\end{figure}

Fig.~\ref{fig:zero-one-generalization} shows the classification performance of ten randomly selected robots in held-out test images, which were originally evolved to recognize a particular image of zero and one. We see that the test performance of the robots varies a lot, sometimes dropping below random guessing performance. 
These robots that achieve less than random guessing performances are the ones that fail to locomote more than three voxel lengths for some images. Although they may have moved in the right direction, we consider these images as misclassified, since the robot fails to achieve its primary goal of locomotion.
On the other hand, seven out of ten robots show test performance greater than 80\%, and the average test accuracy of ten randomly chosen robots is 76.57\%, demonstrating that evolution has discovered robust ways to recognize each digit through training on only one example of each digit. The ability to robustly handle variations in input suggests that the robots are forming a connection between visual data and motion, rather than solely optimizing a trajectory.

\begin{figure}[t]
    \centering
    \includegraphics[width=\linewidth]{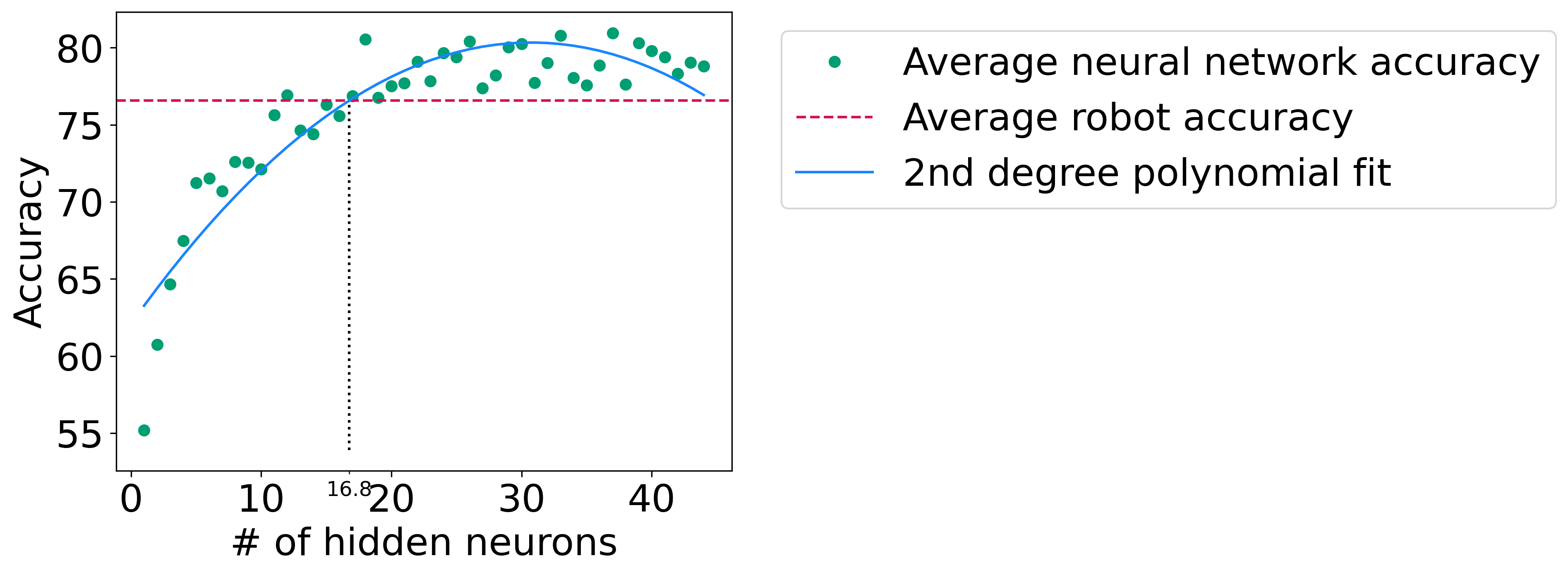}
    \caption{Our attempt to quantify morphological cognition in robots evolved to recognize digits zero and one. Each dot shows the test accuracy of a single-layer MLP with varying numbers of hidden neurons, averaged over 50 training runs from scratch with different random seeds. The solid line shows the 2\textsuperscript{nd} degree polynomial fitted to the data. The dashed line shows the average test performance of 10 randomly selected robots. The robot's performance is roughly equal to a single-layer MLP with 16.8 hidden neurons. }
    \label{fig:nn-vs-robot-zero-one}
\end{figure}

\begin{table}
    \centering
    \begin{tabular}{|c|c|c|}
        \hline
        \textbf{Behavior} & \textbf{Test Accuracy} &
        \begin{tabular}[c]{@{}c@{}} \textbf{Morphological}  \textbf{Cognition} \\ \textbf{Capacity} (MLP -- CNN)\end{tabular}
        \\ \hline
        0 / 1      & 76.57\%  & MLP-16.8 -- CNN-5.9    \\ \hline
        0 / 1,2,3  & 68.47\%  & No Match -- No Match   \\ \hline
        1 / 0,2,3  & 58.96\%  & No Match -- CNN-16.2   \\ \hline
        2 / 0,1,3  & 70.01\%  & No Match -- No Match  \\ \hline
        3 / 0,1,2  & 72.00\%  & No Match -- No Match  \\ \hline
        0,1 / 2,3  & 58.22\%  & No Match -- CNN-15.1  \\ \hline
        0,2 / 1,3  & 62.02\%  & No Match -- CNN-24.0  \\ \hline
        0,3 / 1,2  & 57.14\%  & No Match -- CNN-13.7  \\ \hline
    \end{tabular}
    \caption{
    Test performance of 10 randomly selected robots for each evolved behavior and their morphological cognition capacities. Each robot is evaluated on 1000 held-out images. Their morphological cognition capacities are approximated in terms of neural networks with similar generalization performance. No match refers to the cases where no experimented neural network matches the generalization performance of the robots.
    }
    \label{tab:mce}
\end{table}

Lastly, we attempt to measure the morphological cognition capacity in evolved robots. Our approach is based on a comparison with neural networks, which are commonly utilized to tackle such classification tasks.
We train fully connected single hidden layer neural networks (MLPs) with varying numbers of neurons in the hidden layer ($\in [1, 45]$), and train convolutional neural networks (CNNs) with a single convolutional layer and an output head, with varying numbers of kernels in the convolutional layer  ($\in [1, 45]$). 
We compare the networks' test performance with the robots' test performance.
To keep the comparison fair in terms of the data being used, we train the neural networks with the same data the robot fitnesses are evaluated with during evolution -- a single example from each experimented class. Training is performed with Adam optimizer for 50 epochs or until early stopping with patience five calculated on the test set\footnote{This is not a good practice for the purpose of evaluating machine learning methods. However, we use it here to calculate the best-case performance for the baseline that we are comparing against.}. 

Fig.\ref{fig:nn-vs-robot-zero-one} compares the average test performance of 10 randomly selected robots to MLPs of varying sizes, to illustrate the process. The dashed line marks the average test performance of the robots. Each dot shows the average test performance of the MLPs. 
The solid line shows the 2\textsuperscript{nd} degree polynomial fitted to this data. The polynomial crosses the robot performance for the first time at approximately 16 hidden neurons, indicating that the morphological cognition exhibited by these robots is roughly equivalent to a single-layer MLP with 16 hidden neurons.

We perform an identical analysis on robots that exhibit every evolved behavior -- we select 10 random robots for each possible behavior from our two-digit and four-digit experiments. Table~\ref{tab:mce} summarizes the results. We see that the performance of the robots varies depending on the task challenge. Moreover, we find that in none of the 4-digit classification tasks, MLPs achieve the same level of generalization performance. In four of seven four-digit classification tasks, we discover CNNs that achieve similar generalization performance, effectively quantifying the morphological cognition capacity of the selected robots. Interestingly, in three out of seven tasks, no experimented neural network was able to achieve the same level of generalization performance -- robots that consist of simple materials and no abstract controller achieve better generalization performance than neural networks that have been tested in learning from a single example. 

\section{Conclusion}

In this work, we experiment with evolving simulated voxel-based soft robots to classify MNIST digits. The robots are evolved to move in different directions in the presence of images of different classes. Importantly, we show that this type of behavior can be achieved without an abstract controller. Instead, the robot's behavior emerges from the behaviors of its voxels, where each voxel has a fixed and simple behavior. 
Specifically, we show that by combining two types of voxels with simple shape-changing behaviors, namely muscle voxels that continuously actuate and sensor voxels that can expand or contract proportionally to a pixel intensity, we can evolve robots that carry out a cognitive task of classification. 
Moreover, these robots show a remarkable ability to generalize from a single example, matching, or in some cases exceeding, the performance of simple fully connected or convolutional neural networks. We refer to these robots' ability to perform cognitive tasks without any controller as ``morphological cognition''.

We hope that this work provides an important, and hopefully entertaining, proof-of-concept demonstration that high-level cognition can emerge from simple morphological processes. In future work, we hope to study different morphological processes and how they can give rise to different cognitive behaviors. Lastly, we fully acknowledge that most complex and adaptive behavior in nature is not fully morphological, but is a result of an interplay between the brain, the body, and the environment. We hope to investigate the interactions between different types of intelligence and the implications of such combinations in terms of robustness, adaptability, and optimization.

\section{Acknowledgements}
This material is based upon work supported by the National Science Foundation under Grant No. 2239691 and 2218063.  Thanks to the Vermont Advanced Computing Center for providing computational resources.

The authors thank Neil Traft for his feedback and coining the term ``morphological cognition''.

\footnotesize
\bibliographystyle{apalike}
\bibliography{example} 

\end{document}